\documentclass[letterpaper, 10 pt, conference]{ieeeconf}  %

\IEEEoverridecommandlockouts                              %

\usepackage{microtype}
\let\labelindent\relax
\usepackage{enumitem}
\usepackage{url}
\usepackage{hyperref}

\title{\LARGE \bf
Quantile Transfer for Reliable Operating Point Selection in\\Visual Place Recognition
}

\author{Dhyey Manish Rajani \quad Michael Milford \quad Tobias Fischer
\thanks{This research was supported by the QUT Centre for Robotics, and funding from ARC Laureate Fellowship FL210100156 to MM, as well as an ARC DECRA Fellowship DE240100149 to TF.}
\thanks{The authors are with the QUT Centre for Robotics, School of Electrical
Engineering and Robotics, Queensland University of Technology, Brisbane,
QLD 4000, Australia. Email:
        {\tt\small dhyeymanish.rajani@hdr.qut.edu.au}}%
}

\usepackage{amsmath, amssymb}
\usepackage{booktabs} %
\usepackage{multirow}  %
\usepackage{graphicx}
\usepackage[capitalise]{cleveref}
\usepackage{xcolor} 
\definecolor{darkgreen}{RGB}{0,90,0}  %
\usepackage{cite}

\usepackage{eso-pic}
\usepackage{url}
\AddToShipoutPicture*{%
     \AtTextUpperLeft{%
         \put(-3.5,10){
           \begin{minipage}{\textwidth}
              \scriptsize
              \MakeUppercase{Preprint version of an IEEE/RSJ International Conference on Intelligent Robots and Systems (IROS) 2026 paper; final version to appear on IEEE Xplore}
           \end{minipage}}%
         }%
}

\begin{document}
\bstctlcite{IEEEexample:BSTcontrol}

\maketitle
\thispagestyle{empty}
\pagestyle{empty}

\begin{abstract}
Visual Place Recognition (VPR) is a key component for localization in Global Navigation Satellite System (GNSS)-denied environments, but its performance critically depends on selecting an image matching threshold (operating point) that balances precision and recall. 
Thresholds are typically hand-tuned offline for a specific environment and fixed during deployment, leading to degraded performance under environmental change.
We propose a method that automatically estimates the operating point of a VPR system to maximize recall whilst aiming to achieve 100\% precision. 
The method uses a small calibration traversal with known correspondences and transfers thresholds to deployment via quantile normalization of similarity score distributions. 
This quantile transfer ensures that thresholds remain stable across calibration sizes and query subsets.
Experiments with seven state-of-the-art VPR techniques across five benchmark datasets demonstrate that our proposed approach consistently outperforms existing baselines, enabling the underlying VPR technique to operate at 100\% precision in approximately twice as many deployment scenarios (median improvement), while retrieving up to 29\% more correct matches at that precision.
The method eliminates manual tuning by adapting to new environments and generalizing across operating conditions. 
Our code is available at~\href{https://github.com/DhyeyR-007/Quantile-Transfer-for-Reliable-VPR}{https://github.com/DhyeyR-007/Quantile-Transfer-for-Reliable-VPR}.

\end{abstract}

\section{Introduction}

Visual Place Recognition (VPR) enables mobile robots to localize in GNSS-denied environments by recognizing previously visited locations and supporting loop closure in Visual Simultaneous Localization and Mapping (V-SLAM) frameworks \cite{Zhang2021VisualPerspective,Masone2021ARecognition,Yin2025GeneralAutonomy,ijcai2021p603,schubert2023visual}. VPR is typically formulated as an image retrieval problem: descriptors are extracted from query and database images, and their similarity, commonly measured via cosine similarity, determines potential matches.

\begin{figure}
    \centering
    \includegraphics[width=0.90\columnwidth]{New_figures_IROS2026/draft3_1-cover_figure_IROS2026.pdf}
    \caption{\textbf{Overview of the proposed quantile transfer method for operating point selection in Visual Place Recognition (VPR).} A calibration traversal with known correspondences is used to estimate thresholds that satisfy a 100\% precision requirement. These thresholds are converted into quantiles of the calibration score distribution and transferred to deployment traversals, yielding operating thresholds that maximize recall while satisfying the 100\% precision requirement, without requiring ground truth labels online.}
    \label{fig:cover_figure}
\end{figure}

A key decision is whether to accept or reject a retrieved image as a place match. If the similarity exceeds a matching threshold (MT), the system accepts the match (a true positive if correct, or a false positive otherwise); if it falls below, the match is rejected (a true negative if correct, or a false negative otherwise). The threshold thus controls the precision--recall trade-off: raising it reduces false positives but sacrifices recall, while lowering it admits more true matches but increases false positives.

Applications are widespread: while robust optimization in visual SLAM can handle some false positive loop closures, an excessive number can inundate the optimizer, compromising its online operation. Similarly, in visual localization, geometric verification is computationally expensive and cannot be applied to all candidates. %
Controlling false positives at the VPR retrieval stage therefore directly reduces the number of candidates forwarded to such expensive geometric processing, which is infeasible to apply exhaustively at scale.

To achieve a high-precision operating point, such as 100\% precision, threshold selection is often performed manually \emph{offline} by plotting precision–recall curves against ground truth correspondences~\cite{schubert2021makes}. 
The resulting  threshold is then statically applied at deployment time. 
However, dynamically estimating the MT could adapt to changes in viewpoint, appearance, or perceptual aliasing. 

In this paper, we address this gap by proposing a method that automatically and dynamically estimates a MT that maximizes recall at a target precision level, with a particular focus on perfect (100\%) precision, relevant to safety- or operationally-critical scenarios where even a single false positive can be catastrophic, but maximizing recall is still important.
We emphasize that the objective of our proposed method is not to improve the retrieval performance of a VPR system per se~\cite{schubert2023visual, ijcai2021p603}, but rather to substantially increase the likelihood of the system operating at a high precision point whilst maximizing recall. 

In extensive experiments across seven state-of-the-art VPR techniques and five datasets spanning diverse environmental conditions including seasonal, illumination, appearance, and viewpoint changes, 
we demonstrate consistent and substantial improvements over the state-of-the-art in both automatically deploying the VPR system in a way that achieves 100\% precision \textit{and} maximizes recall at 100\% precision. 
Further analysis provides characterization of the system's behavior including the required amount of calibration data. 

\section{Related Work}

In this section, we review related work on visual place recognition (VPR) and operating parameter selection in VPR.

\subsection{Visual Place Recognition (VPR)}

VPR underpins vision-based localization and V-SLAM loop closure, and has been comprehensively surveyed in recent reviews \cite{Zhang2021VisualPerspective,Masone2021ARecognition,Yin2025GeneralAutonomy,ijcai2021p603,schubert2023visual,zaffar2021vpr}. VPR is typically cast as an image retrieval task, where the query image's feature descriptor is compared to the database descriptors by computing a similarity metric such as the cosine similarity. A central theme in VPR research entails developing expressive image descriptors robust to perceptual aliasing, appearance, and viewpoint changes \cite{schubert2023visual,zaffar2021vpr, zaffar2024estimation,Zhang2021VisualPerspective,Masone2021ARecognition}. Early VPR techniques relied on hand-crafted features \cite{lowry2015visual, zaffar2021vpr}, while modern methods employ deep neural networks to extract more robust image descriptors \cite{berton2025megaloc, izquierdo2024optimal, ali2024boq, lu2024cricavpr, ali2023mixvpr, berton2022rethinking, arandjelovic2016netvlad,Zhang2021VisualPerspective,Masone2021ARecognition}.

Alongside descriptor design, a growing line of research investigates introspection to enable a VPR system to ``know when it doesn’t know'' \cite{carson2022predicting}. Introspection methods assess the reliability of their predictions. Recent approaches estimate localization integrity in a single-image VPR framework, filtering out unreliable matches via supervised learning \cite{carson2022predicting} and later extended to unsupervised settings \cite{10160679}. A similar approach demonstrated introspection applied to actively navigating robots~\cite{claxton2024improving}. While our approach is not strictly an introspection method, it shares the post-hoc use of similarity scores to regulate retrieval behavior, positioning it as a complementary strategy for improving reliability.

The notion of introspection extends beyond visual localization or VPR to system-level integrity or reliability in autonomous robotics, quantifying the trustworthiness of a system’s actions for a given task \cite{zhu2022integrity}. It is widely applied in safety-critical and online operation \cite{hafez2020quantifying, arana2020localization, pietrantuono2018robotics}, enabling systems to either anticipate possible outcomes \cite{guruau2018learn} or reactively identify and correct errors \cite{gautam2022method,arana2020localization, zhu2022integrity}, underscoring the broad relevance of introspection across autonomous robotics.

\subsection{Selecting Visual Place Recognition Operating Parameters}
\label{subsec:operatingparameters}

The operating parameters of a VPR system, such as image resolution, coverage, or similarity thresholds, strongly influence downstream performance \cite{mount2020unsupervised, schubert2023visual}. Prior work has shown that modifying input resolution impacts robustness under appearance change \cite{milford2012seqslam}, while calibration routines can be used to set parameters that trade coverage against recall in surface-based localization \cite{mount2019automatic}. Building on this, \cite{mount2020unsupervised} proposed an unsupervised method to automatically tune operating configurations, validated across diverse VPR datasets \cite{sunderhauf2013we, maddern20171}.
Calibration has also been applied to the matching scores themselves. Gronát et al.~\cite{gronat2016learning} learn a separate classifier per database location and calibrate their outputs to make scores comparable across locations. We share the premise that raw scores are not directly comparable, but transfer an image matching threshold across environments rather than learning a model per location.

Operating a VPR system at a fixed high-precision point and reporting the recall attained there has been a standard practice in visual localization. For instance, Chen et al.~\cite{chen2011city} evaluate city-scale landmark retrieval in terms of recall at 95\% precision. Yet selecting an image matching threshold that achieves such an operating point, despite directly determining the precision–recall trade-off, has received comparatively limited attention.
Schubert et al.~\cite{schubert2021beyond} model similarity scores with a robust normal distribution assuming that most query–database pairs are non-matches, and set a fixed high quantile as threshold. Vysotska et al.~\cite{vysotska2025adaptive} extend this idea to sequential VPR by fitting Gaussian mixtures to recent scores, enabling adaptive thresholding under appearance changes.

Our work is most closely related to these approaches. 
In contrast to~\cite{vysotska2025adaptive}, which operates on sequences, we focus on single-image VPR. %
Compared to the unsupervised heuristic of~\cite{schubert2021beyond}, our method uses calibration data to satisfy the 100\% precision requirement while remaining agnostic to the underlying VPR model.

\section{Methodology} \label{methodology_section}

Our goal is to estimate an image matching threshold (MT) that maximizes recall subject to a precision requirement of 100\%. Our method assumes access to similarity scores produced by a VPR technique, along with a small calibration subset of queries with known ground truth correspondences. Using this limited calibration, we infer thresholds for unseen deployment queries, without requiring their ground truth.  

\noindent The procedure consists of four stages (see \Cref{fig:methodology}):  
\begin{enumerate}[label=\Alph*)]
    \item \textbf{Similarity matrices:} construct similarity matrices from calibration and deployment queries.  
    \item \textbf{Adaptive calibration sampling:} identify calibration queries whose score distributions most resemble those of the deployment queries.  
    \item \textbf{Quantile transfer:} convert thresholds learned on calibration into quantiles, which transfer robustly across distributions and yield deployment thresholds.  
    \item \textbf{Operating point selection:} map a user-defined precision requirement to a deployment threshold that maximizes recall while satisfying the constraint.  
\end{enumerate}

\subsection{Similarity Matrices and Calibration Data}  
Let $DB$ denote the database traversal and $Q$ the set of query images. A similarity matrix $S \in \mathbb{R}^{|DB|\times|Q|}$ is computed using cosine similarity between descriptors. The query set is partitioned into a set of calibration queries $\{Q^{cal}\}$ with ground truth correspondences, and deployment (or validation) queries $\{Q^{val}\}$ without ground truth. This yields $S^{cal}$ and $S^{val}$, with binary ground truth labels $GT^{cal}$ available only for calibration.

\subsection{Adaptive Calibration Sampling}  
Thresholds derived directly from $S^{cal}$ may not generalize to $S^{val}$ because similarity distributions differ in different parts of the environment. To address this, we identify calibration queries most similar to each deployment query in terms of their \textit{ranked similarity distributions}.  

For each validation query $j$, let $v_j \in \mathbb{R}^{|DB|}$ denote the $j$-th column of $S^{val}$, i.e. the similarity scores between that query and all $|DB|$ database images. Similarly, let $c_i \in \mathbb{R}^{|DB|}$ be the $i$-th column of $S^{cal}$. After ranking database images by similarity, yielding $v'_j, c'_i \in \mathbb{R}^{|DB|}$, we compute the Pearson correlation between $v'_j$ and every $c'_i$. This yields a correlation matrix $R \in \mathbb{R}^{|Q^{val}|\times|Q^{cal}|}$.
For each $v_j$, we select the $k$ most correlated calibration queries, forming adapted calibration subsets $\{C_k^*\}$ with their associated ground truth labels $\{GT_k^*\}$, such that $C_k^* \in \mathbb{R}^{|DB|\times|Q^{val}|}$ and $GT_k^* \in \mathbb{R}^{|DB|\times|Q^{val}|}$.

\begin{figure}[t]
    \centering
    \includegraphics[width=\columnwidth]{New_figures_IROS2026/2-methodology_IROS2026.pdf}

    \caption{
    \textbf{Methodology of our proposed quantile transfer approach.} A VPR technique is first applied to generate similarity matrices from a database traversal and both calibration and deployment query sets (blue box). Each deployment query is then matched to its most similar calibration queries using a correlation matrix, yielding adapted subsets of calibration similarity matrices (green box). From these subsets, thresholds that maximize recall under a 100\% precision requirement are computed and expressed as quantiles of the calibration score distributions (yellow box). The weighted median value of these quantiles is then transferred to the deployment similarity distribution to obtain the matching threshold without requiring ground truth labels during deployment (pink box).
    }

    \label{fig:methodology}
\end{figure}

\subsection{Quantile-based Threshold Transfer} \label{quantile_thresh_transfer}
The key insight is that while absolute similarity scores vary across environments due to descriptor scaling and environmental factors, the relative ordering of similarities is more stable. By expressing thresholds as quantiles, i.e.~positions within the similarity distribution, we capture this relative structure and enable robust transfer across different environments.

For each adapted calibration subset $C_k^*$, we compute the precision–recall curve using $GT_k^*$ and extract the matching threshold $\theta$ that maximizes recall subject to a predefined precision $\tau \in [0,1]$. Concretely, for each subset $C_k^*$ we evaluate candidate thresholds and retain the one (i.e., $\theta_k$) with the highest recall while ensuring that the achieved precision $\tau^* \geq \tau$ (such that $\tau$ = 1.0 or 100\%).

Since absolute thresholds vary across calibration and deployment distributions, we instead express each subset's threshold $\theta_k$ as a \textit{quantile} $q_k$ within its calibration distribution:
$q_k = \Pr \big[ C_k^* \leq \theta_k \big]$.
Here, $\Pr$ denotes the empirical probability (i.e. the fraction of scores). Intuitively, this converts an absolute similarity threshold into a relative position within the calibration distribution, making it transferable to other distributions with different scales.

The quantiles $\{q_k\}$ then serve as transferable parameters from the calibration set to the validation set. To transfer thresholds to deployment, we apply a quantile $q_k$ to the deployment similarity distribution $S^{val}$. The empirical quantile function $F_{S^{val}}^{-1}$ is obtained by sorting all similarity scores in $S^{val}$ and interpolating if necessary. Given a quantile $q_k$, the deployment threshold is then 
\begin{equation}
\theta^{val}_k = F_{S^{val}}^{-1}(q_k),
\label{eq:eq_2}
\end{equation}
which corresponds to the similarity score at the $q_k$-th quantile of $S^{val}$. This procedure ensures that thresholds retain their relative position in the score distribution when transferred from calibration to deployment.

\subsection{Operating Point Selection}

The final stage establishes a direct mapping from a precision requirement to an operating threshold on the deployment set. During calibration, each quantile $q_k$ is associated with a precision-recall curve, yielding a \iffalse an oracle\fi mapping from quantiles to achievable precision levels.

At deployment, given a predefined precision of $\tau=100\%$, we take the weighted median of the quantile values $\{q_k\}$ obtained above, where quantiles derived from more correlated calibration subsets are weighted more. This yields a single representative quantile, which is then mapped to the corresponding deployment threshold $\theta^{val}$ via \Cref{eq:eq_2}. This threshold is then applied to deployment queries, aiming to achieve recall close to the maximum possible under the precision constraint $\tau$.

\section{Experimental Setup}

To evaluate the effectiveness and generality of our proposed method, we conduct experiments across a diverse set of VPR techniques (Section~\ref{vpr_techniques}), baseline approaches (Section~\ref{baseline_techniques}), and benchmark datasets (Section~\ref{datasets_section}). %
Performance is assessed using complementary metrics that measure how reliably the 100\% precision requirement is satisfied and how much recall is achieved relative to the maximum possible recall at 100\% precision (Section~\ref{Performance_Metrics}).

\subsection{VPR Techniques} \label{vpr_techniques}

We evaluate our method using seven state-of-the-art VPR techniques that span different architectural approaches and descriptor designs: MegaLoc \cite{berton2025megaloc}, SALAD \cite{izquierdo2024optimal}, BoQ \cite{ali2024boq}, EigenPlaces~\cite{berton2023eigenplaces}, MixVPR \cite{ali2023mixvpr}, CosPlace \cite{berton2022rethinking}, and the widely adopted NetVLAD \cite{arandjelovic2016netvlad}. This diverse selection enables assessment of our method's robustness across different descriptor architectures and its competitiveness against established unsupervised baselines~\cite{schubert2021beyond}. All techniques use cosine similarity for image matching, ensuring consistent similarity score interpretation across methods.

\subsection{Baseline Techniques} \label{baseline_techniques}

We compare against the unsupervised thresholding method of Schubert et al.~\cite{schubert2021beyond}, which models query–database similarity scores with a robust normal distribution, assuming non-matching pairs dominate. A fixed high quantile of the fitted distribution is then used as the matching threshold.

Although both methods rely on similarity quantiles, they differ fundamentally in their objective and estimation approach. Schubert et al.~derive a single threshold estimate from distributional assumptions, primarily targeting very high precision, whereas our approach uses a small calibration set to estimate the similarity quantile that achieves maximum recall at a 100\% precision level, and transfers this threshold to deployment data without imposing parametric assumptions.

\begin{table}[t]
\centering
\footnotesize
\setlength{\tabcolsep}{5pt}
\renewcommand{\arraystretch}{1.15}
\caption{For each dataset, we report the number of calibration set queries (first 10\% of the queries), deployment set queries (remaining 90\% of the queries), segment size (set to ${\approx} 1\%$ of the total number of queries in the dataset), and the number of non-overlapping segments formed from the deployment set for evaluation.}
\label{tab:dataset_specs}
\begin{tabular}{lcccc}
\toprule
\textbf{Dataset} & \textbf{Calib. set} & \textbf{Deploy. set} & \textbf{Seg.} & \textbf{\#} \\
& \textbf{queries} & \textbf{queries} & \textbf{size} & \textbf{Segs.} \\
\midrule
SFU Mountain     & 38   & 347  & 4   & 86 \\
Oxford RobotCar  & 387  & 3489 & 39  & 89 \\
Pittsburgh30k    & 681  & 6135 & 69  & 88 \\
Nordland    & 276  & 2484 & 28  & 88 \\
MSLS val         & 1108 & 9976 & 111 & 89 \\
\bottomrule
\end{tabular}

\end{table}

For fairness, we also implement a calibration-aware variant of~\cite{schubert2021beyond}. Here, the same calibration data used by our method is provided to the baseline, which applies robust normal fitting on this data and transfers the resulting threshold to deployment similarities. This preserves the baseline’s estimation procedure while controlling for differences in available information.

\begin{table*}[t]
\caption{Performance comparison of our proposed method against baselines (Schubert et al.~\cite{schubert2021beyond} and Schubert et al. (w/ calib.)) across seven VPR techniques and five datasets, in terms of \%Segs and AvgR@100P ($\tau = 1.0$). \%Segs denotes the percentage of deployment segments in which the estimated MT achieves 100\% precision, and AvgR@100P denotes the average recall achieved across all deployment segments. Bold indicates the best result and underlining indicates the second-best result in each column. Higher values are better for both metrics.}
\label{tab:table_1}
\centering
\large
\setlength{\tabcolsep}{5pt}
\renewcommand{\arraystretch}{1.15}
\resizebox{\textwidth}{!}{%
\begin{tabular}{lll cc cc cc cc cc cc cc |  cc}

\toprule 

& \multicolumn{2}{c}{\textbf{Methods}}
& \multicolumn{2}{c}{\textbf{MegaLoc }} & \multicolumn{2}{c}{\textbf{SALAD }} & \multicolumn{2}{c}{\textbf{BoQ }} & \multicolumn{2}{c}{\textbf{EigenPlaces }} & \multicolumn{2}{c}{\textbf{MixVPR }} & \multicolumn{2}{c}{\textbf{CosPlace }} & \multicolumn{2}{c}{\textbf{NetVLAD }} & \multicolumn{2}{c}{\textbf{Mean}} \\
\cmidrule(lr){4-5}\cmidrule(lr){6-7}\cmidrule(lr){8-9}\cmidrule(lr){10-11}
\cmidrule(lr){12-13}\cmidrule(lr){14-15}\cmidrule(lr){16-17}\cmidrule(lr){18-19}
& & & \shortstack{\%\\Segs} & \shortstack{Avg\\R@100P} & \shortstack{\%\\Segs} & \shortstack{Avg\\R@100P} & \shortstack{\%\\Segs} & \shortstack{Avg\\R@100P} & \shortstack{\%\\Segs} & \shortstack{Avg\\R@100P} & \shortstack{\%\\Segs} & \shortstack{Avg\\R@100P} & \shortstack{\%\\Segs} & \shortstack{Avg\\R@100P} & \shortstack{\%\\Segs} & \shortstack{Avg\\R@100P} & \shortstack{\%\\Segs} & \shortstack{Avg\\R@100P} \\

\midrule  %

\multirow{3}{*}{\textit{\begin{tabular}[c]{@{}l@{}}SFU Mountain \\ (DB: Dry; Q: Dusk) \end{tabular}}}

& & Schubert et al.~\cite{schubert2021beyond} & \shortstack{69.8} & \textbf{\shortstack{0.61}} & \underline{\shortstack{19.8}} & \underline{\shortstack{0.15}} & \underline{\shortstack{14.0}} & \underline{\shortstack{0.08}} & \underline{\shortstack{29.1}} & \underline{\shortstack{0.17}} & \underline{\shortstack{31.4}} & \underline{\shortstack{0.22}} & \underline{\shortstack{17.4}} & \underline{\shortstack{0.08}} & \underline{\shortstack{24.4}} & \underline{\shortstack{0.09}} & \underline{29.4} & \underline{0.20} \\
& & Schubert et al. (w/ calib.) & \underline{\shortstack{72.1}} & \shortstack{0.50} & \shortstack{0.0} & \shortstack{0.00} & \shortstack{0.0} & \shortstack{0.00} & \shortstack{0.0} & \shortstack{0.00} & \shortstack{9.3} & \shortstack{0.03} & \shortstack{0.0} & \shortstack{0.00} & \shortstack{17.4} & \shortstack{0.06} & \shortstack{14.1} & \shortstack{0.08} \\
& & Ours & \textbf{\shortstack{87.2}} & \underline{\shortstack{0.51}} & \textbf{\shortstack{86.0}} & \textbf{\shortstack{0.54}} & \textbf{\shortstack{75.6}} & \textbf{\shortstack{0.47}} & \textbf{\shortstack{81.4}} & \textbf{\shortstack{0.56}} & \textbf{\shortstack{79.1}} & \textbf{\shortstack{0.60}} & \textbf{\shortstack{75.6}} & \textbf{\shortstack{0.47}} & \textbf{\shortstack{52.3}} & \textbf{\shortstack{0.19}} & \textbf{76.7} & \textbf{0.48} \\

\midrule  %

\multirow{3}{*}{\textit{\begin{tabular}[c]{@{}l@{}}Oxford RobotCar \\ (DB: Rain; Q: Dusk) \end{tabular}}}

& & Schubert et al.~\cite{schubert2021beyond} & \shortstack{23.6} & \textbf{\shortstack{0.22}} & \underline{\shortstack{23.6}} & \underline{\shortstack{0.23}} & \textbf{\shortstack{43.8}} & \textbf{\shortstack{0.17}} & \underline{\shortstack{23.6}} & \underline{\shortstack{0.11}} & \shortstack{22.5} & \underline{\shortstack{0.22}} & \underline{\shortstack{11.2}} & \underline{\shortstack{0.08}} & \shortstack{7.9} & \shortstack{0.03} & \underline{22.3} & \underline{0.15} \\
& & Schubert et al. (w/ calib.) & \underline{\shortstack{24.7}} & \textbf{0.22} & \underline{\shortstack{23.6}} & \underline{\shortstack{0.23}} & \underline{\shortstack{31.5}} & \underline{\shortstack{0.08}} & \shortstack{14.6} & \shortstack{0.08} & \underline{\shortstack{23.6}} & \underline{\shortstack{0.22}} & \shortstack{7.9} & \underline{\shortstack{0.08}} & \underline{\shortstack{27.0}} & \textbf{\shortstack{0.07}} & \shortstack{21.8} & \shortstack{0.14} \\
& & Ours & \textbf{\shortstack{34.8}} & \textbf{\textbf{0.22}} & \textbf{\shortstack{29.2}} & \textbf{\shortstack{0.24}} & \shortstack{30.3} & \textbf{\shortstack{0.17}} & \textbf{\shortstack{42.7}} & \textbf{\shortstack{0.14}} & \textbf{\shortstack{30.3}} & \textbf{\shortstack{0.26}} & \textbf{\shortstack{37.1}} & \textbf{\shortstack{0.13}} & \textbf{\shortstack{40.4}} & \underline{\shortstack{0.05}} & \textbf{35.0} & \textbf{0.18} \\

\midrule  %

\multirow{3}{*}{\textit{\begin{tabular}[c]{@{}l@{}}Pittsburgh30k \end{tabular}}}

& & Schubert et al.~\cite{schubert2021beyond} & \underline{\shortstack{44.3}} & \underline{\shortstack{0.44}} & \underline{\shortstack{28.4}} & \underline{\shortstack{0.28}} & \underline{\shortstack{28.4}} & \underline{\shortstack{0.28}} & \underline{\shortstack{30.7}} & \underline{\shortstack{0.31}} & \underline{\shortstack{31.8}} & \underline{\shortstack{0.32}} & \underline{\shortstack{27.3}} & \underline{\shortstack{0.27}} & \shortstack{15.9} & \shortstack{0.16} & \shortstack{29.5} & \underline{0.30} \\
& & Schubert et al. (w/ calib.) & \underline{\shortstack{44.3}} & \underline{\shortstack{0.44}} & \underline{\shortstack{28.4}} & \underline{\shortstack{0.28}} & \underline{\shortstack{28.4}} & \underline{\shortstack{0.28}} & \underline{\shortstack{30.7}} & \underline{\shortstack{0.31}} & \underline{\shortstack{31.8}} & \underline{\shortstack{0.32}} & \underline{\shortstack{27.3}} & \underline{\shortstack{0.27}} & \underline{\shortstack{18.2}} & \underline{\shortstack{0.18}} & \underline{29.9} & \underline{0.30} \\
& & Ours & \textbf{\shortstack{61.4}} & \textbf{\shortstack{0.51}} & \textbf{\shortstack{39.8}} & \textbf{\shortstack{0.35}} & \textbf{\shortstack{59.1}} & \textbf{\shortstack{0.30}} & \textbf{\shortstack{59.1}} & \textbf{\shortstack{0.34}} & \textbf{\shortstack{55.7}} & \textbf{\shortstack{0.35}} & \textbf{\shortstack{61.4}} & \textbf{\shortstack{0.28}} & \textbf{\shortstack{76.1}} & \textbf{\shortstack{0.20}} & \textbf{58.9} & \textbf{0.33} \\

\midrule  %

\multirow{3}{*}{\textit{\begin{tabular}[c]{@{}l@{}}Nordland \\ (DB: Winter; Q: Summer)\end{tabular}}}

& & Schubert et al.~\cite{schubert2021beyond} & \underline{\shortstack{30.7}} & \underline{\shortstack{0.30}} & \textbf{\shortstack{36.4}} & \textbf{\shortstack{0.21}} & \shortstack{0.0} & \underline{\shortstack{0.00}} & \textbf{\shortstack{17.0}} & \underline{\shortstack{0.04}} & \underline{\shortstack{1.1}} & \underline{\shortstack{0.01}} & \underline{\shortstack{13.6}} & \underline{\shortstack{0.02}} & \underline{\shortstack{0.0}} & \textbf{\shortstack{0.00}} & \underline{14.1} & \underline{0.08} \\
& & Schubert et al. (w/ calib.) & \shortstack{28.4} & \shortstack{0.28} & \shortstack{15.9} & \shortstack{0.16} & \underline{\shortstack{2.3}} & \underline{\shortstack{0.00}} & \shortstack{2.3} & \shortstack{0.02} & \underline{\shortstack{1.1}} & \underline{\shortstack{0.01}} & \shortstack{1.1} & \shortstack{0.01} & \underline{\shortstack{0.0}} & \textbf{\shortstack{0.00}} & \shortstack{7.3} & \shortstack{0.07} \\
& & Ours & \textbf{\shortstack{44.3}} & \textbf{\shortstack{0.40}} & \underline{\shortstack{22.7}} & \underline{\shortstack{0.19}} & \textbf{\shortstack{6.8}} & \textbf{\shortstack{0.01}} & \underline{\shortstack{13.6}} & \textbf{\shortstack{0.05}} & \textbf{\shortstack{28.4}} & \textbf{\shortstack{0.09}} & \textbf{\shortstack{19.3}} & \textbf{\shortstack{0.05}} & \textbf{\shortstack{3.4}} & \textbf{\shortstack{0.00}} & \textbf{19.8} & \textbf{0.11} \\

\midrule  %

\multirow{3}{*}{\textit{\begin{tabular}[c]{@{}l@{}}Nordland \\ (DB: Fall; Q: Summer) \end{tabular}}}

& & Schubert et al.~\cite{schubert2021beyond} & \underline{\shortstack{60.2}} & \underline{\shortstack{0.60}} & \underline{\shortstack{65.9}} & \underline{\shortstack{0.64}} & \underline{\shortstack{60.2}} & \underline{\shortstack{0.19}} & \textbf{\shortstack{64.8}} & \underline{\shortstack{0.51}} & \textbf{\shortstack{67.0}} & \textbf{\shortstack{0.60}} & \textbf{\shortstack{55.7}} & \shortstack{0.38} & \underline{\shortstack{2.3}} & \shortstack{0.00} & \textbf{53.7} & \underline{0.42} \\
& & Schubert et al. (w/ calib.) & \shortstack{59.1} & \shortstack{0.59} & \shortstack{60.2} & \shortstack{0.60} & \textbf{\shortstack{79.5}} & \shortstack{0.18} & \shortstack{52.3} & \shortstack{0.50} & \shortstack{53.4} & \shortstack{0.52} & \underline{\shortstack{46.6}} & \textbf{\shortstack{0.43}} & \underline{\shortstack{2.3}} & \underline{\shortstack{0.01}} & \shortstack{50.5} & \shortstack{0.41} \\
& & Ours & \textbf{\shortstack{67.0}} & \textbf{\shortstack{0.63}} & \textbf{\shortstack{70.5}} & \textbf{\shortstack{0.66}} & \shortstack{34.1} & \textbf{\shortstack{0.32}} & \underline{\shortstack{59.1}} & \textbf{\shortstack{0.55}} & \underline{\shortstack{60.2}} & \underline{\shortstack{0.56}} & \shortstack{45.5} & \underline{\shortstack{0.42}} & \textbf{\shortstack{27.3}} & \textbf{\shortstack{0.06}} & \underline{51.9} & \textbf{0.46} \\

\midrule %

\multirow{3}{*}{\textit{\begin{tabular}[c]{@{}l@{}} MSLS val \end{tabular}}}

& & Schubert et al.~\cite{schubert2021beyond} & \underline{\shortstack{12.4}} & \underline{\shortstack{0.12}} & \underline{\shortstack{3.4}} & \underline{\shortstack{0.03}} & \shortstack{0.0} & \shortstack{0.00} & \shortstack{5.6} & \shortstack{0.06} & \underline{\shortstack{1.1}} & \underline{\shortstack{0.01}} & \shortstack{3.4} & \shortstack{0.03} & \shortstack{0.0} & \underline{\shortstack{0.00}} & \shortstack{3.7} & \underline{\shortstack{0.04}} \\
& & Schubert et al. (w/ calib.) & \underline{\shortstack{12.4}} & \underline{\shortstack{0.12}} & \underline{\shortstack{3.4}} & \underline{\shortstack{0.03}} & \underline{\shortstack{1.1}} & \underline{\shortstack{0.01}} & \underline{\shortstack{7.9}} & \underline{\shortstack{0.08}} & \underline{\shortstack{1.1}} & \underline{\shortstack{0.01}} & \underline{\shortstack{4.5}} & \underline{\shortstack{0.04}} & \underline{\shortstack{1.1}} & \underline{\shortstack{0.00}} & \underline{4.5} & \underline{0.04} \\ 
& & Ours & \textbf{\shortstack{58.4}} & \textbf{\shortstack{0.33}} & \textbf{\shortstack{40.4}} & \textbf{\shortstack{0.19}} & \textbf{\shortstack{52.8}} & \textbf{\shortstack{0.12}} & \textbf{\shortstack{52.8}} & \textbf{\shortstack{0.24}} & \textbf{\shortstack{48.3}} & \textbf{\shortstack{0.20}} & \textbf{\shortstack{50.6}} & \textbf{\shortstack{0.23}} & \textbf{\shortstack{68.5}} & \textbf{\shortstack{0.06}} & \textbf{53.1} & \textbf{0.19} \\

\bottomrule

\end{tabular}%
}
\end{table*}

\subsection{Datasets} 
\label{datasets_section}

We evaluate on five standard VPR benchmarks covering viewpoint, appearance, seasonal and illumination variation, across urban and rural environments:

\begin{enumerate}
    \item \textbf{Nordland~\cite{sunderhauf2013we}:} We use the Winter–Summer and Fall–Summer traverses, following~\cite{zaffar2021vpr, izquierdo2024optimal}. We follow the sampling in~\cite{zaffar2021vpr}, yielding 27,592 database and 2,760 query images. The ground truth tolerance is $\pm$1 frame.

    \item \textbf{SFU Mountain~\cite{bruce2015sfu}:} Following~\cite{hussaini2025improving}, we use Dry (DB) vs.~Dusk (query) with 385 images per traverse and set the ground truth tolerance to $\pm$0 frames. SFU was captured in a semi-structured environment with challenging illumination and appearance conditions.

    \item \textbf{Oxford RobotCar~\cite{maddern20171}:} Following~\cite{hussaini2025improving}, we use the Rain (DB) vs.~Dusk (query) traverses captured in an urban environment with varying weather conditions, with 3,876 images per traverse. The ground truth tolerance is $\pm$5 frames.
    
    \item \textbf{Pittsburgh30k~\cite{arandjelovic2016netvlad}:} We use the standard Pittsburgh30k-test split, following~\cite{berton2022deep}. This leads to 10,000 database and 6,816 query images captured in different years under substantial viewpoint and appearance variations, with a ground truth tolerance of $\pm$25 m.

    \item \textbf{MSLS~\cite{warburg2020mapillary}:} We use the validation subset covering multiple cities over several years, following~\cite{warburg2020mapillary, berton2022deep}. This leads to 18,871 database and 11,084 query images, with a ground truth tolerance of $\pm$25 m.
    
\end{enumerate}

The experimental specifications for each VPR dataset used in \Cref{results_section}, including the number of calibration queries, deployment queries, segment size, and total number of segments, are specified in Table~\ref{tab:dataset_specs}.

\subsection{Performance Metrics} \label{Performance_Metrics}

To evaluate performance, a VPR dataset is first divided into calibration and deployment sets. The first 10\% of the queries, for which ground truth is available, are used for calibrating our method. The remaining queries form the deployment set, ensuring temporal separation and no overlap with the calibration data. %
We partition the deployment set into non-overlapping segments of fixed size. %
For each segment, a MT is estimated.

We compute performance over two measures:
\begin{itemize} %

    \item \textbf{\% Segments with Precision (P) $\geq \tau$ (\%Segs):} The primary objective of our method is to select a MT that ensures the VPR system operates at a predetermined precision level $\tau$. Hence, \%Segs is used to measure the proportion of deployment set segments in which the estimated MT successfully achieves P $\geq \tau$. Higher \%Segs indicates that the estimated MT consistently enables the underlying VPR system to operate at the precision constraint $\tau$. %
    In our experiments, we set $\tau$ = 1.0 (or 100\% precision).

    \item \textbf{Average Recall achieved at $\tau$ Precision (AvgR@$\tau$P):} While \%Segs indicates how often the precision requirement $\tau$ is met across the deployment set segments, it does not reflect the number of correct matches retrieved at that precision. Trivially, a very high similarity score set as MT, being very conservative, could satisfy the precision constraint across all segments but yield very few correct matches, resulting in low recall. Hence, AvgR@$\tau$P is used to quantify the average recall achieved across all segments when using the estimated MT, where segments in which the estimated MT fails to meet the precision constraint $\tau$ contribute zero recall.
    Higher AvgR@$\tau$P provides a measure of the system’s retrieval effectiveness at the specified precision level $\tau$.
    In our experiments, this corresponds to AvgR@100P (i.e., $\tau$ = 100\% precision).

\end{itemize}

\section{Results}
\label{results_section}

We evaluate our quantile transfer method against the baselines across seven VPR techniques and five datasets. Performance is reported in terms of \%Segs and AvgR@100P.

\subsection{Overall Performance}

Table~\ref{tab:table_1} summarizes results across all methods and datasets. Our approach consistently outperforms both Schubert et al.~\cite{schubert2021beyond} and its calibration variant in \%Segs and AvgR@100P across the majority of VPR techniques and datasets.

The joint improvement in both metrics indicates that our thresholds are neither overly conservative (high \%Segs, low recall) nor overly liberal (low \%Segs due to precision violations). Instead, our method more reliably satisfies the 100\% precision constraint while maximizing recall at that operating point.

Gains are particularly pronounced on MSLS val and SFU Mountain, where both baseline methods frequently fail to meet the precision constraint. On Pittsburgh30k and Oxford RobotCar, improvements are consistent but more moderate. On Nordland Winter–Summer, all methods struggle due to severe seasonal change; nevertheless, our method achieves the highest mean performance. On Fall–Summer, Schubert et al. attains slightly higher \%Segs on average, while our method achieves higher AvgR@100P, indicating a less conservative but still reliable threshold.

Across all datasets, the calibration variant of Schubert et al.~offers no systematic benefit over the original baseline, suggesting that applying fixed distributional assumptions to calibration data does not generalize across deployment segments. In contrast, our method leverages calibration data to adapt thresholds to deployment conditions, resulting in more stable precision control and higher recall.

\subsection{Dataset-Level Trends}

To illustrate Table~\ref{tab:table_1}, consider MegaLoc on Nordland (DB: Winter; Q: Summer). The first 10\% of queries (276 images) are used for calibration, leaving 2,484 deployment queries divided into 88 non-overlapping segments of 28 queries. Our method achieves 100\% precision in 39 segments (\%Segs = 44.3) and an average recall of 40\% across all segments (AvgR@100P = 0.40), compared with 27 segments and 30\% recall for Schubert et al., and 25 segments and 28\% recall for Schubert et al.~(w/ calib.).

On SFU Mountain, our method achieves the highest mean \%Segs and AvgR@100P. Although the baselines sometimes attain higher recall in segments satisfying the precision constraint, they do so less consistently and therefore achieve lower overall \%Segs. The calibration variant also collapses to zero \%Segs for several VPR techniques, whereas our method remains strong.

On Oxford RobotCar, AvgR@100P is similar across methods, but our approach substantially improves \%Segs, demonstrating more reliable precision enforcement under challenging illumination changes. On Pittsburgh30k, the two Schubert variants perform nearly identically, indicating that calibration does not alter their fixed threshold estimate, whereas our method substantially increases \%Segs while maintaining comparable recall.

On Nordland, Schubert et al. achieves higher \%Segs in six cases: SALAD and EigenPlaces on Winter–Summer, and BoQ, EigenPlaces, MixVPR, and CosPlace on Fall–Summer. However, \%Segs can be increased trivially using a conservative threshold that accepts few positives and therefore produces few false positives, motivating the complementary AvgR@100P metric. In four of these six cases, our method retrieves more correct matches despite its lower \%Segs: BoQ (0.32 vs.\ 0.19), EigenPlaces (0.55 vs.\ 0.51), and CosPlace (0.42 vs.\ 0.38) on Fall–Summer, and EigenPlaces (0.05 vs.\ 0.04) on Winter–Summer. For example, on BoQ (Fall–Summer), Schubert et al.~satisfies the precision constraint in 60.2\% of segments but achieves only 0.19 recall, whereas our method achieves 0.32 recall over 34.1\% of segments. Thus, the baseline satisfies the constraint more often only by sacrificing matches needed by downstream systems.

Our method is genuinely outperformed on both metrics in the remaining two cases: SALAD on Winter–Summer (22.7 vs.\ 36.4 \%Segs; 0.19 vs.\ 0.21 recall) and MixVPR on Fall–Summer (60.2 vs.\ 67.0 \%Segs; 0.56 vs.\ 0.60 recall). Because our method targets the maximum-recall operating point at 100\% precision, it estimates a threshold near the precision boundary and is more sensitive to estimation error than a conservative baseline when true and false matches are poorly separated. In these cases, the transferred quantile yields a slightly permissive threshold that admits false positives, reducing \%Segs; because segments violating the precision constraint contribute zero recall, AvgR@100P also declines.

As a supervised method, our approach benefits most when the calibration traverse represents deployment, as on most datasets and reflected in its higher mean performance. Gains narrow when a VPR technique poorly separates true and false matches, as for SALAD on Winter–Summer and MixVPR on Fall–Summer. Nevertheless, calibration requirements are modest: the ablation in~\Cref{subsec:ablation_calib_set_size} shows that only 1\% of the dataset is sufficient.

On MSLS, our method substantially increases \%Segs relative to both baselines, whose performance remains near zero, highlighting the benefit of adaptive threshold transfer under strong distributional variability.

\begin{figure}[t]
    \centering
    \includegraphics[width=\columnwidth]{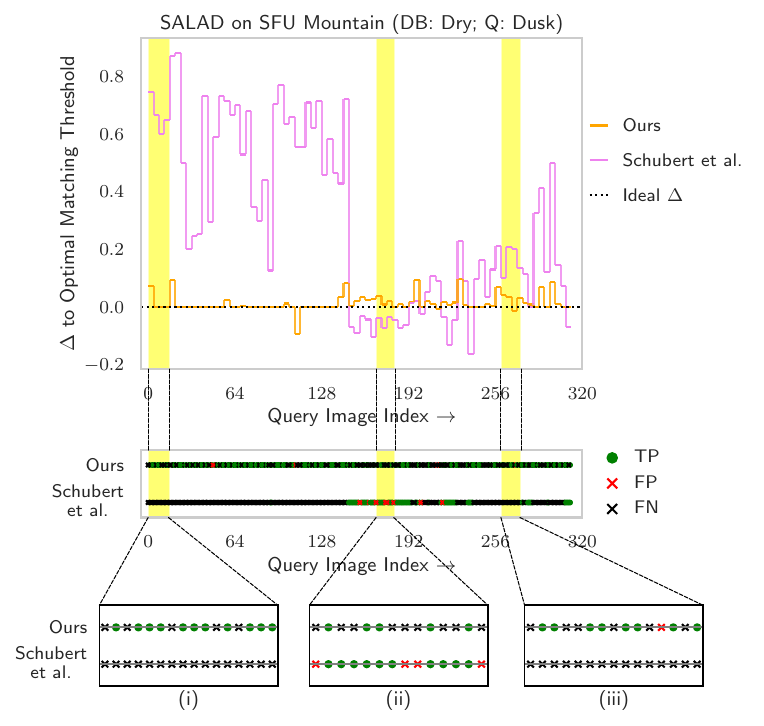}
    \caption{
    \textbf{Dynamic Matching Threshold (MT) estimation for SALAD on SFU Mountain (DB: Dry; Q: Dusk).} 
    \textit{Top:} The deviation ($\Delta$) of the estimated MT from the optimal MT (i.e., the threshold achieving maximum recall at 100\% precision, computed using ground truth), across queries, where values closer to zero indicate more accurate threshold estimation. {\color{orange} Our method's} MT estimate remains consistently close to the ideal $\Delta$ = 0 (dotted line), while {\color{magenta} Schubert et al.} overestimates the threshold for the majority of deployment queries, deviating significantly from the optimal across the traversal. 
    \textit{Bottom:} {\color{darkgreen} True Positives (TP)}, {\color{red} False Positives (FP)}, and False Negatives (FN) decisions for each query image obtained using {\color{orange} Our method} and {\color{magenta} Schubert et al.'s} MT estimates. 
    Three representative portions (i), (ii), and (iii) from the traversal are zoomed in to illustrate qualitatively the difference in retrievals between the two methods.
    }
    \label{fig:matchingthresholds}
    \vspace{-0.3cm}
\end{figure}

\subsection{Adaptive Matching Threshold}

Figure~\ref{fig:matchingthresholds} compares threshold estimates for SALAD on SFU Mountain. Our method tracks the segment-wise optimal threshold closely (MSE = 0.010), whereas Schubert et al.~exhibits substantially larger deviations (MSE = 0.182).

This translates directly to retrieval behavior: in the case shown in Figure~\ref{fig:matchingthresholds}, across the traversal, our method yields 193 true positives, 4 false positives, and 115 false negatives, compared to 76 true positives, 9 false positives, and 227 false negatives for Schubert et al., directly illustrating that our estimated MT not only retrieves far more true positives but also reduces false positives. 
Deviations from the optimal threshold are smaller and more consistent, explaining the higher \%Segs and AvgR@100P observed in Table~\ref{tab:table_1}.

\subsection{Consistency Across VPR Backbones}

Beyond average performance, an important observation from Table~\ref{tab:table_1} is the consistency of improvements across heterogeneous VPR backbones, including global descriptor methods (e.g., NetVLAD-style), transformer-based models, and lightweight embeddings. Our method improves mean \%Segs on the majority of backbones for every dataset, rather than relying on gains from a single strong model.

This indicates that the benefit arises from threshold estimation itself rather than from interactions with a specific similarity distribution. In contrast, the Schubert variants exhibit large variance across backbones, particularly under strong distribution shift (e.g., MSLS val), where performance collapses for several models. The stability of our gains suggests that quantile transfer generalizes across fundamentally different similarity score statistics.

\subsection{Qualitative Results}

\begin{figure}[t]
    \centering
    \includegraphics[width=\columnwidth]{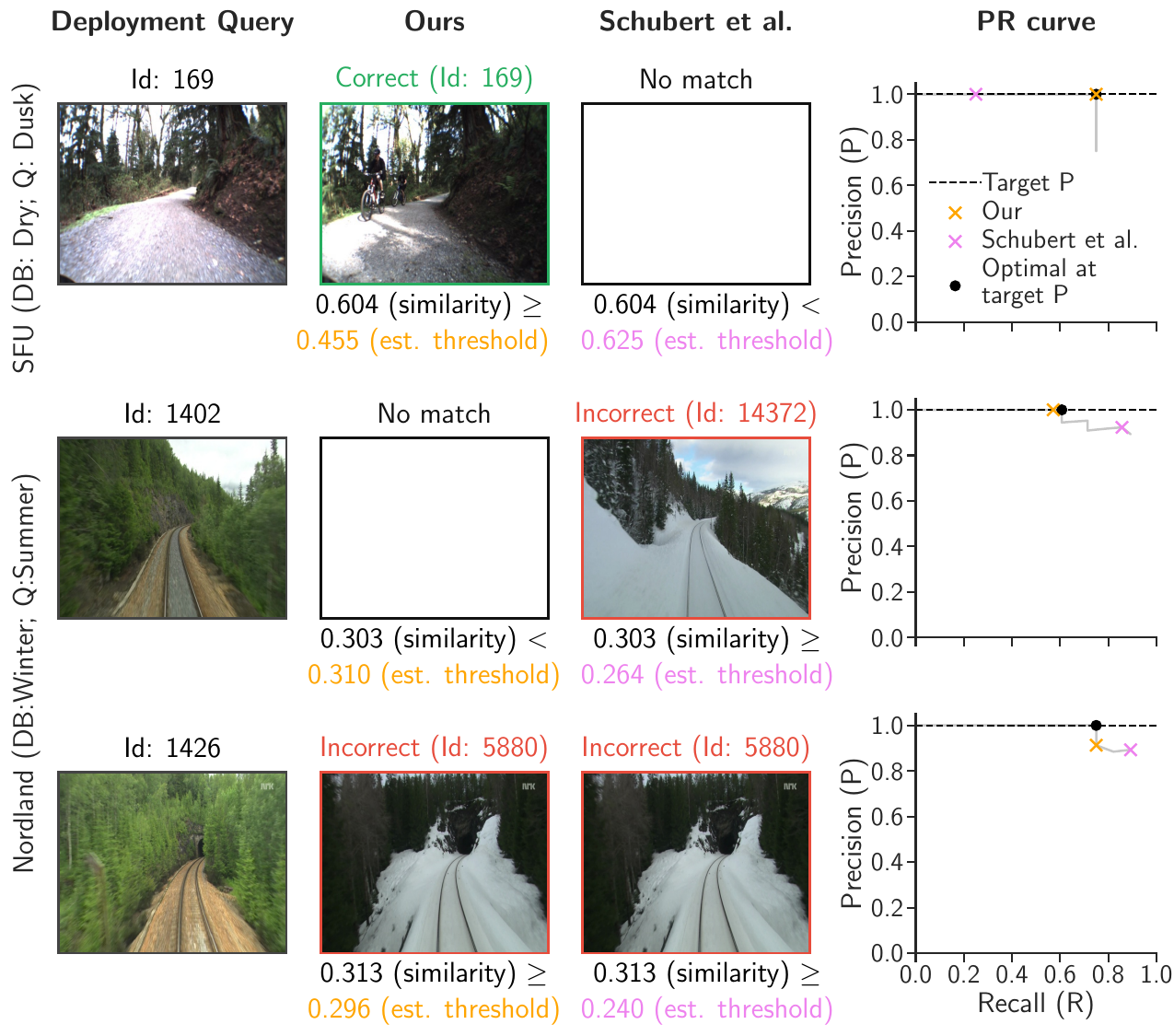}
    \caption{
    \textbf{Qualitative examples comparing matching decisions by our method and Schubert et al. baseline across Nordland and SFU Mountain datasets}. For each query image (\textit{left}), we show the best-matching database image retrieved under each method's estimated Matching Threshold (MT) (\textit{middle}), with {\color[HTML]{27ae60}{green}} and {\color[HTML]{e74c3c}{red}} borders indicating correct (True Positives) and incorrect matches (False Positives); and empty boxes indicating no match retrieved (False Negatives). The associated Precision--Recall (PR) curves (\textit{right}) plot the MT estimated by {\color{orange} Our method ($\times$)}, {\color{magenta} Schubert et al. ($\times$)}, and the optimal MT (i.e., the threshold that actually achieves maximum recall at 100\% precision)  ($\bullet$).
    {\color{orange} Our method} accepts true positives that {\color{magenta} Schubert et al.} incorrectly rejects (\textit{top row}), while simultaneously filtering out false positives that {\color{magenta} Schubert et al.} would admit (\textit{middle row}), though in some cases both methods fail (\textit{bottom row}) with our estimate being marginally worse.
    These examples illustrate that our approach achieves a better balance between recall and precision, adapting thresholds to accept informative matches without sacrificing reliability.
    }

    \label{fig:qualitative}
    \vspace{-0.4cm}
\end{figure}

Figure~\ref{fig:qualitative} illustrates representative examples. On SFU Mountain, our method recovers true matches rejected by the baseline due to overly conservative thresholds. On Nordland, it avoids false positives admitted by overly liberal baseline thresholds. Failure cases occur for both methods, but our estimates remain closer to the optimal operating point. These examples directly reflect the quantitative improvements under strict precision constraints.

\subsection{Ablation: Calibration Set Size} \label{subsec:ablation_calib_set_size}

\begin{figure}[t]
    \centering
    \includegraphics[width=\columnwidth]{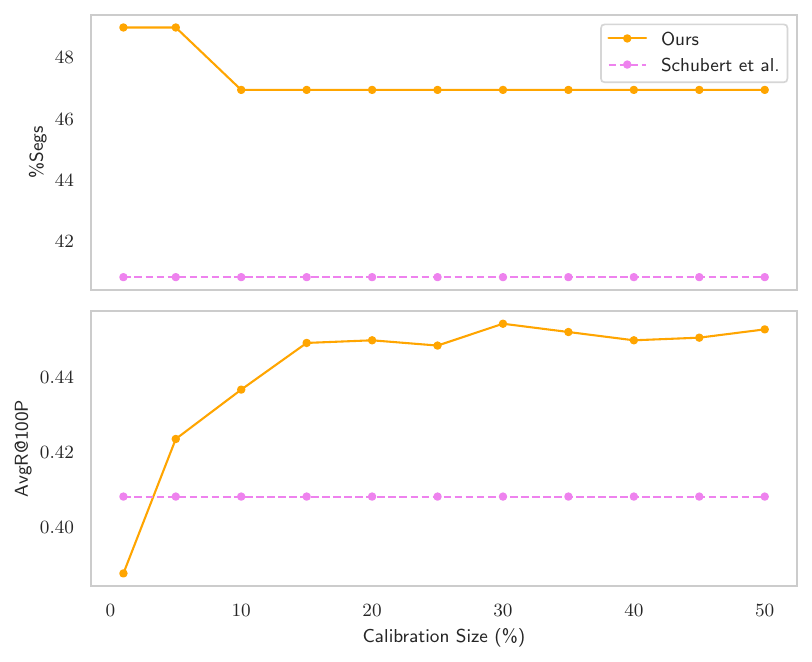}
    \vspace*{-0.1cm}

    \caption{
    \textbf{Ablation study of our method on calibration set size using MegaLoc on Nordland (DB: Fall; Q: Summer).} The calibration proportion is varied from 1\% to 50\% of the dataset, with the evaluation set fixed to the remaining 50\%. {\color{orange} Our method} consistently outperforms {\color{magenta} Schubert et al.}~baseline in \%Segs (\textit{top}) across all the calibration sizes, and improves in AvgR@100P (\textit{bottom}) with more calibration data. This demonstrates that {\color{orange} Our method} does not require a large, labeled calibration set representative of deployment to operate effectively.
    }

    \label{fig:ablation_figure}
    \vspace{-0.4cm}
\end{figure}

Figure~\ref{fig:ablation_figure} evaluates sensitivity to calibration set size (1–50\%), while keeping the deployment set fixed to the last 50\% of the dataset. Our method consistently outperforms Schubert et al. in \%Segs across all calibration sizes, demonstrating that even a very small labeled calibration set (as low as 1\% of the dataset) is sufficient for reliable operation at 100\% precision. AvgR@100P improves gradually with larger calibration sets before plateauing around 30\%, indicating better quantile estimation with broader coverage. %
 
Schubert et al. remains unchanged across all calibration sizes as its threshold derives from fixed distributional assumptions, unaffected by the availability of labeled calibration data. These results demonstrate that our MT estimation approach remains effective even with limited calibration data.

\subsection{Runtime}

On an Intel i7-14700 CPU, calibration sampling and quantile transfer require approximately 0.0202s per query on Nordland, 0.0005s on SFU Mountain, 0.0030s on Oxford RobotCar, 0.0138s on Pittsburgh30k, and 0.0514s on MSLS val, using vectorized NumPy operations and multiprocessing across available CPU cores.

\section{Conclusions}
\label{sec:conclusions}

We presented a quantile transfer method that automatically selects operating thresholds for visual place recognition systems that maximize recall at 100\% precision. Our work is agnostic to the rapidly improving field of descriptor and front-end matching development, and provides a mechanism to maximize recall whilst achieving a 100\% precision operating point. Extensive evaluation across seven state-of-the-art VPR methods and five benchmark datasets demonstrates substantial performance gains over the baseline at 100\% precision.

There are several promising areas for future research. First, the proposed method has primarily been demonstrated on instantaneous, single-frame matching techniques; extending it to sequential or filtering-based methods would further increase its generality. The current system relies on a static calibration set, which could potentially be updated online through self-supervision, especially as the environment changes over time. A further extension could eliminate the need for an initial static calibration set entirely, instead curating one online through self-supervision, particularly when local correspondence cues such as self-motion are available.

\bibliographystyle{IEEEtran}    %
\bibliography{references}       %

\end{document}